\documentclass[twoside,11pt]{article}

\usepackage{jmlr2e}
\usepackage{listings}

\jmlrheading{x}{2017}{x}{3/17}{x}{x}{Olson, La Cava, Orzechowski, Urbanowicz, and Moore}

\ShortHeadings{PMLB: A Large Benchmark Suite for Machine Learning Evaluation and Comparison}{Olson, La Cava, Orzechowski, Urbanowicz, and Moore}

\firstpageno{1}

\begin{document}

\title{PMLB: A Large Benchmark Suite for Machine Learning Evaluation and Comparison}

\author{\name Randal S. Olson \email olsonran@upenn.edu \\
       \AND
       \name William La Cava \email lacava@upenn.edu \\
       \AND
       \name Patryk Orzechowski \email patryk@upenn.edu \\ 
       \AND
       \name Ryan J. Urbanowicz \email ryanurb@upenn.edu \\
       \AND
       \name Jason H. Moore \email jhmoore@upenn.edu \\
       \addr Institute for Biomedical Informatics\\
        University of Pennsylvania\\
        3700 Hamilton Walk, Philadelphia, PA 19104, USA
}

\editor{[empty for now]}

\maketitle

\begin{abstract}
The selection, development, or comparison of machine learning methods in data mining can be a difficult task based on the target problem and goals of a particular study.  Numerous publicly available real-world and simulated benchmark datasets have emerged from different sources, but their organization and adoption as standards have been inconsistent.  As such, selecting and curating specific benchmarks remains an unnecessary burden on machine learning practitioners and data scientists. The present study introduces an accessible, curated, and developing public benchmark resource to facilitate identification of the strengths and weaknesses of different machine learning methodologies. We compare meta-features among the current set of benchmark datasets in this resource to characterize the diversity of available data.  Finally, we apply a number of established machine learning methods to the entire benchmark suite and analyze how datasets and algorithms cluster in terms of performance.  This work is an important first step towards understanding the limitations of popular benchmarking suites and developing a resource that connects existing benchmarking standards to more diverse and efficient standards in the future.
\end{abstract}

\begin{keywords}
machine learning, benchmarking, data repository, classification
\end{keywords}

\section{Introduction}
The term \emph{benchmarking} is used in machine learning (ML) to refer to the evaluation and comparison of ML methods regarding their ability to learn patterns in `benchmark' datasets that have been applied as `standards'.  Benchmarking could be thought of simply as a sanity check to confirm that a new method successfully runs as expected and can reliably find simple patterns that existing methods are known to identify \cite{MachineLearningBook}. A more rigorous way to view benchmarking is as an approach to identify the respective strengths and weaknesses of a given methodology in contrast with others \cite{caruana2006empirical}.  Comparisons could be made over a range of evaluation metrics, e.g., power to detect signal, prediction accuracy, computational complexity, and model interpretability.  This approach to benchmarking would be important for demonstrating new methodological abilities or simply to guide the selection of an appropriate ML method for a given problem.  

Benchmark datasets typically take one of three forms. The first is accessible, well-studied \emph{real-world data}, taken from different real-world problem domains of interest. The second is \emph{simulated data}, or data that has been artificially generated, often to `look' like real-world data, but with known, underlying patterns. For example, the GAMETES genetic-data simulation software generates epistatic patterns of association in `mock' single nucleotide polymorphism (SNP) data \cite{urbanowicz2012gametes,urbanowicz2012predicting}. The third form is \emph{toy data}, which we will define here as data that is also artificially generated with a known embedded pattern but without an emphasis on representing real-world data, e.g., the parity or multiplexer problems \cite{Blum2003,Koza1992}. It is worth noting that the term `toy dataset' has often been used to describe a small and simple dataset such as the examples included with algorithm software.  

While some benchmark repositories and datasets have emerged as more popular than others, ML still lacks a central, comprehensive, and concise set of benchmark datasets that accentuate the strengths and weaknesses of established ML methods.  Individual studies often restrict their benchmarking efforts for various reasons, for example based on comparing variants of the ML algorithm of interest. The genetic programming (GP) community has also previously discussed appropriate benchmarking when comparing GP methodologies \cite{ONeill2010,McDermott2012,White2013}. Benchmarking efforts may focus on a specific application of interest, e.g. traffic sign detection \cite{stallkamp2012man}, or a more narrowly defined ML problem type, e.g. classification of 2-way epistatic interactions \cite{moore2006flexible,li2016detecting}.  The scope of benchmarking may also be limited by practical computational requirements.  

There are currently a number of challenges that make it difficult to benchmark ML methods in a useful and globally accepted manner. For one, there are an overwhelming number of publications that reference the use of benchmark datasets, however there are surprisingly few publications that discuss the topic of appropriate ML benchmarking in general. Additionally, collecting and curating real-world benchmark datasets remains a challenge for many researchers. Although repositories such as the UCI ML repository~\cite{Lichman2013} and Kaggle \cite{kaggle} provide dozens of real-world datasets to download for free, these datasets come in myriad formats and require considerable preprocessing before ML methods can be applied to them. As a result, many benchmark datasets go unused simply because they are too difficult to preprocess. Further, while real-world benchmarks can be derived from many different problem domains, from a strict data science perspective, many of the benchmarks in repositories can have very similar meta-features (e.g. the number of instances, number of features, number of classes, presence of missing data, and similar signal to noise ratios, etc.), such that while they are representative of different real-world problems, they may not represent a diverse assembly of data science problems.  This issue has been raised previously; when applying UCI datasets as benchmarks, it was noted that the scope of included datasets limited method evaluation, and suggested that repositories such as UCI should be expanded \cite{segal2004machine}.

Another challenge in benchmarking is that researchers often use only a handful of datasets when evaluating their methods, which can make it difficult to properly compare one ML method to the state-of-the-art ML methods. For example, these datasets may be handpicked to highlight the strengths of the proposed method, while failing to demonstrate the proposed method's potential weaknesses. As a result, although a ML method may perform well on a handful of datasets, it may fail to generalize to a broader range of problems. We submit that it is just as important to clearly identify the limitations of an algorithm in benchmarking practices, something that is often overlooked.  While there will always be a need to identify and generate custom benchmarks for new or specialized problem domains, e.g. physical activity monitoring data \cite{reiss2012creating} or dynamical systems simulation \cite{la2016inference}, it is vital for the bioinformatics and ML community to have a comprehensive benchmark suite with which to compare and contrast ML methods. Towards this goal, the present study introduces the Penn Machine Learning Benchmark (PMLB), a publicly available dataset suite (accessibly hosted on GitHub) initialized with 165 real-world, simulated, and toy benchmark datasets for evaluating supervised classification methods. PMLB includes datasets from many of the most-used ML benchmark suites, such as KEEL~\cite{Alcala2010} and the UCI ML repository~\cite{Lichman2013}. In addition to collecting data from these resources, PMLB standardizes the format of these data and provides useful interfaces for fetching datasets directly from the web. 

This initial PMLB repository is not meant to be comprehensive; it includes mainly real-world datasets and excludes regression datasets (i.e. those with a continuous-valued dependent variable), as well as any datasets with missing values. We have chosen to focus our initial assessment on available datasets in classification. This paper includes a high-level analysis of the properties (i.e. meta-features) of the founding PMLB datasets, such as feature counts, class imbalance, etc.  Further, we evaluate the performance of 13 standard statistical ML methods from scikit-learn \cite{scikit-learn} over the full set of PMLB datasets. We then assess the diversity of these benchmark datasets from the perspective of their meta-features as well as based on the predictive performance over the set of ML methods applied.  Beyond introducing a new simplified resource for ML benchmarks, this study was designed to provide insight into the limitations of currently utilized benchmarks, and direct the expansion and curation of a future improved PMLB dataset suite that more efficiently and comprehensivly allows for the comparison of ML methods.  This work provides an important first step in the assembly of a effective and diverse set of benchmarking standards integrating real-world, simulated, and toy datasets for generalized ML evaluation and comparison.


\section{Penn Machine Learning Benchmark (PMLB)}

We compiled the Penn Machine Learning Benchmark (PMLB) datasets from a wide range of existing ML benchmark suites including the UCI ML repository \cite{Lichman2013}, Kaggle \cite{kaggle}, KEEL \cite{Alcala2010}, and meta-learning benchmark \cite{Reif2012}.  As such, the PMLB includes most of the real-world benchmark datasets commonly used in ML benchmarking studies. 

To make the PMLB easier to use, we preprocessed all of the datasets to follow a standard row-column format, where the features correspond to columns in the dataset and every instance in the data set is a row. All categorical features and labels with non-numerical encodings were replaced with numerical equivalents (e.g., ``Low", ``Medium", and ``High" were replaced with 0, 1, and 2).  Additionally, in every dataset, the dependent variable column was labeled as ``class''. Finally, all benchmark datasets with missing data were excluded from PMLB, as many ML algorithms cannot handle missing data in their standard implementations and we wished to avoid imposing a particular data imputation method in this initial study.

Currently, the PMLB consists of datasets for supervised classification (binary and multiclass). In supervised classification, we wish to find a mapping ${\hat{y}(\mathbf{x}): \mathbb{R}^p \rightarrow \mathcal{Y}}$ that associates the vector of features $\mathbf{x} \in \mathbb{R}^p$ with  class labels from the set $\mathcal{Y} = \{1\;\dots\;K\}$ using $N$ paired examples from the training set $\mathcal{T} = \{(\mathbf{x}_i,y_i), i = 1\;\dots\;N\}$. In the future we plan to expand PMLB to include datasets for regression.

\subsection{PMLB Meta-Features}

\begin{figure}
    \centering
    \includegraphics[width=0.9\textwidth]{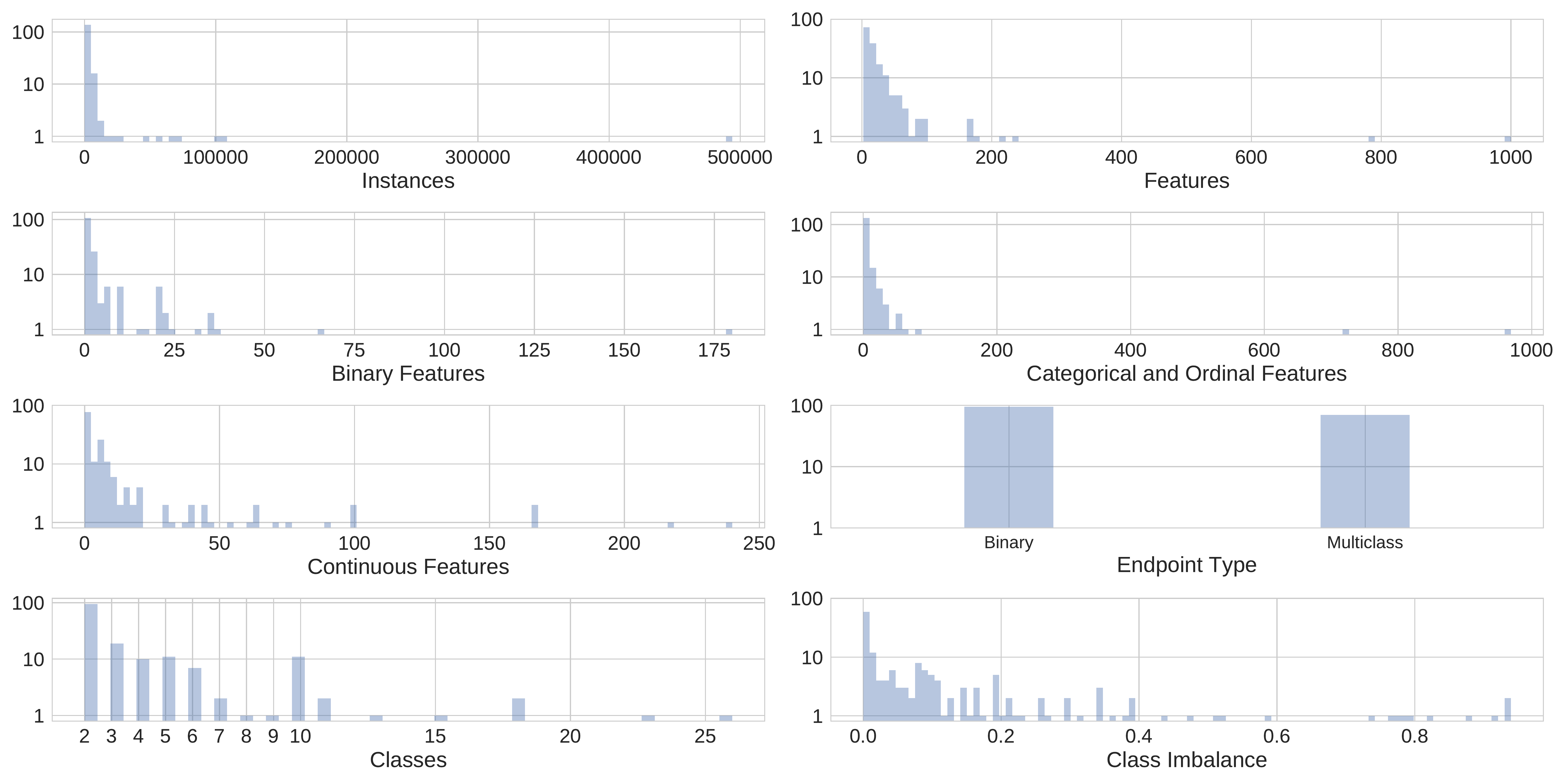}
    \caption{Histograms showing the distribution of meta-feature values from the PMLB datasets. Note the log scale of the y axes.}
    \label{fig:dataset_features}
\end{figure}

In the current release, the PMLB includes 165 datasets.  The meta-features of these datasets are sumarized in Figure~\ref{fig:dataset_features}. These meta-features are defined as follows:

\begin{itemize}
    \item {\bf \# Instances}: The number of instances in each dataset.
    \item {\bf \# Features}: The number of features in each dataset.
    \item {\bf \# Binary Features}: The number of categorical features in each dataset with only two levels.
    \item {\bf \# Categorical and Ordinal Features}: The number of discrete features in each dataset with $>$2 levels. 
    \item {\bf \# Continuous Features}: The number of continuous-valued features in each dataset. Discriminating categorical and ordinal features from continuous features was determined automatically based on whether a variable was considered to be a `float' in a Pandas dataframe \cite{pandas}.
    \item {\bf Endpoint Type}: Whether each dataset is a binary or multiclass supervised classification problem. Again, continuous endpoints for regression have been excluded in this study.
    \item {\bf \# Classes}: The number of classes to predict in each dataset's endpoint.
    \item {\bf Class Imbalance}: The level of class imbalance in each dataset $\in [0 \; 1)$, where 0.0 corresponds to perfectly balanced classes and a value approaching 1.0 corresponds to extreme class imbalance, i.e. where nearly all instances have the same class value. Imbalance is calculated by measuring the squared distance of each class's instance proportion from perfect balance in the dataset, as:
    \[I = K \sum_{i=1}^K{(\frac{n_i}{N} - \frac{1}{K})^2}  \]
    
    where $n_i$ is the number of instances of class $i \in \mathcal{Y} $. 
\end{itemize}

Most of the datasets have under 5,000 instances and 500 features, and a fairly balanced class distribution. Roughly half of the datasets are binary classification problems, whereas the remaining half are multiclass classification problems ranging from 3-26 classes. Of the 165 datasets, 49 datasets have a mix of discrete (i.e. binary, categorical or ordinal) and continuous features, while 12 include only binary features, and 53 contain only continuous features.  It is worth noting that the PMLB datasets cover a broad range of application areas, including biomedical studies, signal processing, and image classification, among others.

\subsection{PMLB Python Interface}

To make the PMLB datasets easier to access, we published an open source Python interface for PMLB on PyPi\footnote{https://pypi.python.org/pypi/pmlb/}. This interface provides a simple \texttt{fetch\_data} function that returns any dataset in the PMLB as a \emph{pandas} \cite{pandas} DataFrame. For example, to fetch the \texttt{clean2} dataset:

\begin{lstlisting}
import pmlb
clean2_data = pmlb.fetch_data(`clean2')
\end{lstlisting}

\noindent The \texttt{clean2\_data} variable will then contain a data frame of the \texttt{clean2} dataset, where the \texttt{class} column corresponds to the class labels and the remaining columns are the features. The \texttt{fetch\_data} function has several caching and preprocessing options, all of which are documented in the PMLB repository\footnote{https://github.com/EpistasisLab/penn-ml-benchmarks}.

To acquire a full list of all datasets available in PMLB, users can access the \texttt{dataset\_names} variable:

\begin{lstlisting}
import pmlb
print(pmlb.dataset_names)
\end{lstlisting}

\noindent which is simply a Python list that contains the names of all PMLB datasets. For the remainder of the experiments described below, we used this Python interface to load the datasets prior to analysis.

\section{Evaluating Machine Learning Methods}

To provide a basis for comparison, we evaluated 13 supervised ML classification methods from scikit-learn~\cite{scikit-learn} on the 165 datasets in PMLB. These methods include:

\begin{itemize}
    \item Gaussian Na\"{i}ve Bayes (NB)
    \item Bernoulli Na\"{i}ve Bayes
    \item Multinomial Na\"{i}ve Bayes
    \item Logistic Regression
    \item Linear classifier trained via stochastic gradient descent (SGD)
    \item Support Vector Classifier (SVC) with a linear, polynomial, sigmoid, or RBF kernel
    \item Passive Aggressive classifier
    \item K-Nearest Neighbor (KNN)
    \item Decision Tree
    \item Random Forest
    \item Extra Random Forest (a.k.a. Extra Trees Classifier)
    \item AdaBoost
    \item Gradient Tree Boosting
\end{itemize}

ML methods were evaluated using {\it balanced accuracy} \cite{Velez2007,urbanowicz2015exstracs} as the scoring metric, which is a normalized version of accuracy that accounts for class imbalance by calculating accuracy on a per-class basis then averaging the per-class accuracies. For more information on these ML methods, see~\cite{MachineLearningBook} and the scikit-learn documentation~\cite{scikit-learn}. When we evaluated each ML method, we first scaled the features of the datasets by subtracting the mean and scaling the features to unit variance. This scaling step was necessary for some ML methods, such as the K-Nearest Neighbor classifier, which assumes that the datasets will be scaled appropriately beforehand. (Note that the datasets provided in PMLB are not scaled or normalized in order to keep them as close as possible to their original form.) 

Once the datasets were scaled, we performed a comprehensive grid search of each of the ML method's parameters using 10-fold cross-validation to find the best parameters (according to mean cross-validation balanced accuracy) for each ML method on each data set. This process resulted in a total of over 5.5 million evaluations of the 13 ML methods over the 165 data sets. For a comprehensive parameter search, we used expert knowledge about the ML methods to decide what parameters and parameter values to evaluate. The complete code for running the experiment is available online\footnote{https://github.com/rhiever/sklearn-benchmarks/tree/master/model\_code/}. It should be noted that due to the different number of parameters for each algorithm, not every algorithm had the same number of evaluations.

\section{Results}

In order to characterize the datasets in PMLB, they are clustered based on their meta-features in Section~\ref{s:r1}. We then analyze the datasets based on ML performance in Section~\ref{s:r2}, which identifies which datasets can be solved with high or low accuracy, as well as which datasets are appear universally easy or hard for the set of different ML algorithms to model accurately versus which ones appear to be particularly useful for highlighting differential ML algorithm performance.

\subsection{Dataset Meta-Features}\label{s:r1}

\begin{figure}
    \centering
    \includegraphics[width=0.9\textwidth]{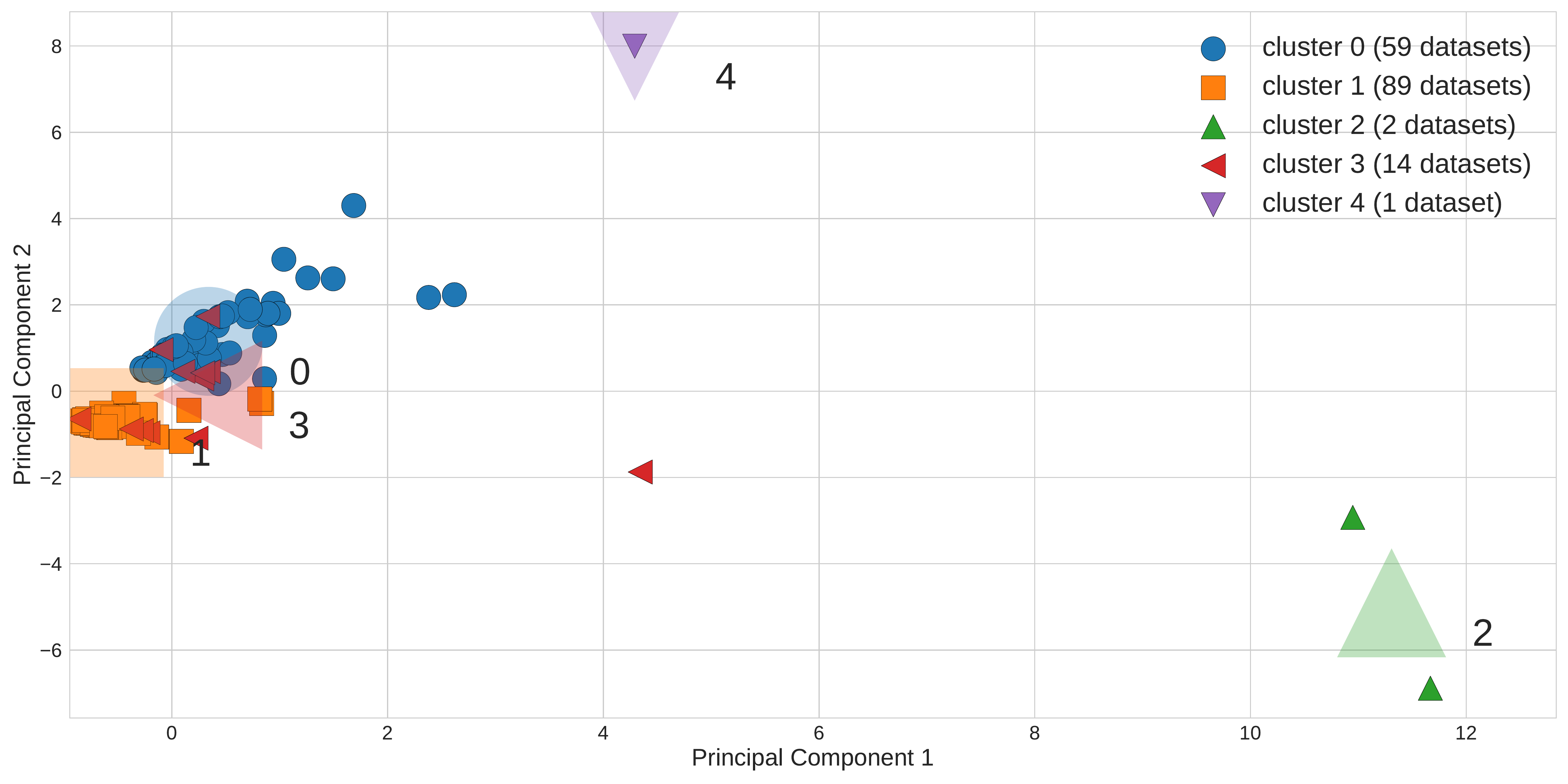}
    \caption{Clustered meta-features of datasets in the PMLB projected onto the first two principal component axes (PCA 1 and PCA 2).}
    \label{fig:pca_data_features}
\end{figure}

\begin{figure}
    \centering
    \includegraphics[width=0.9\textwidth]{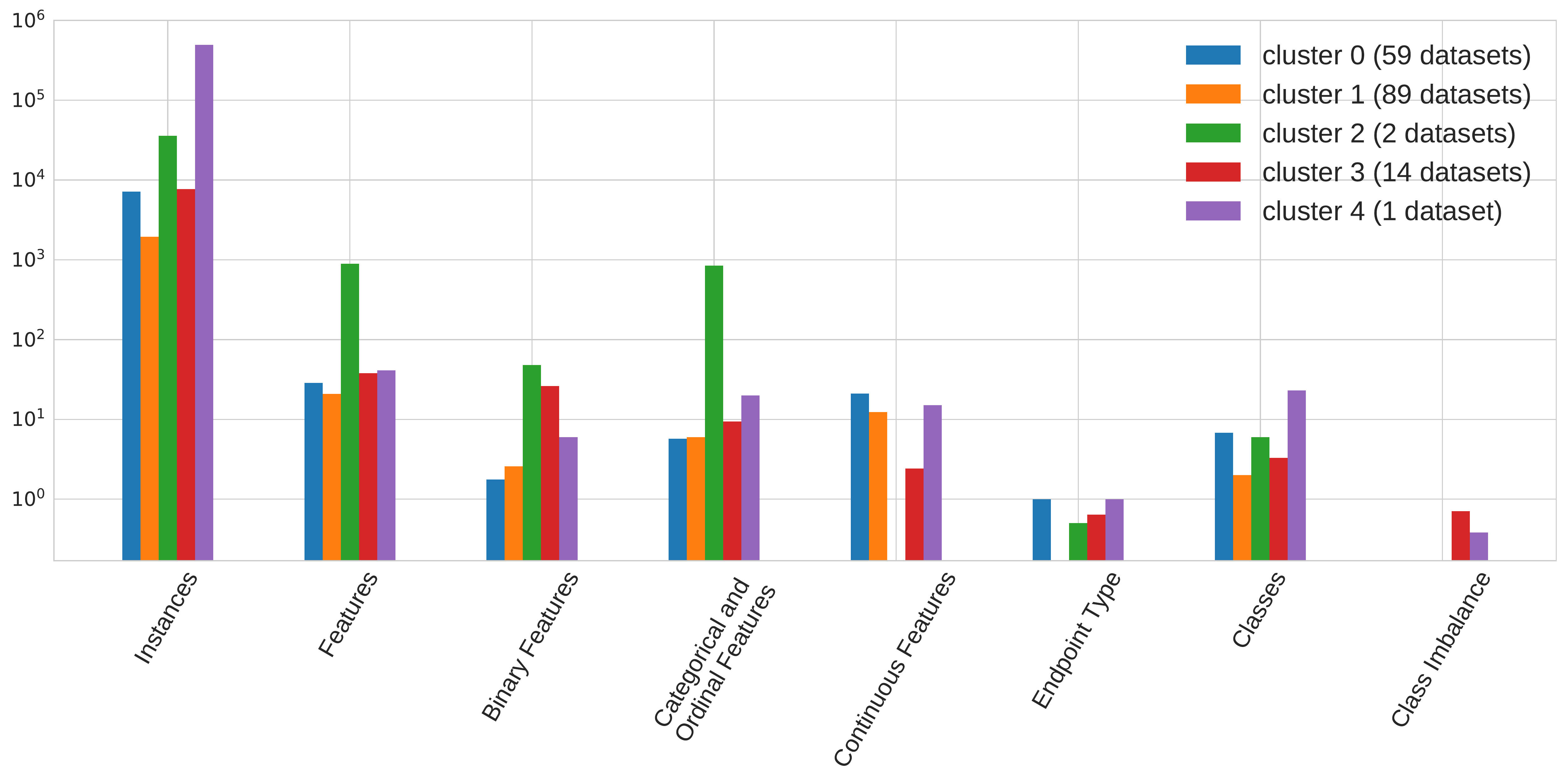}
    \caption{Mean values of each meta-feature within PMLB dataset clusters identified in Figure \ref{fig:pca_data_features}.}
    \label{fig:cluster_dataset_bar}
\end{figure}

We used $k$-means to cluster the normalized meta-features of the datasets into 5 clusters, visualized along the first two principal component axes in Figure~\ref{fig:pca_data_features} (note that the first two components of the PCA explain 49\% of the variance, so we expect there to be some overlap of clusters in visualization). The number of clusters was chosen to compromise between the interpretability of the clusters and the adequate separation of the clustered datasets, as defined by the silhouette score. Figure~\ref{fig:pca_data_features} includes two clusters centered on outlier datasets (clusters 2 and 4). All clusters are compared in more detail according to the mean values of the dataset meta-features in each cluster in Figure~\ref{fig:cluster_dataset_bar}. Clusters 0 and 1 contain most of the datasets, and are separated by their endpoint type, i.e. cluster 0 is comprised of binary classification problems, whereas cluster 1 is comprised of multiclass classification problems. Cluster 2 is made up of 3 datasets with relatively high numbers features (a \texttt{GAMETES} dataset with 1000 features and the \texttt{MNIST} dataset with 784). Cluster 3 contains datasets with high imbalance between classes in the data set. Finally, cluster 4 is reserved for the \texttt{KDD Cup} dataset, which has exceptionally high number of instances (nearly 500,000). The clustering analysis thus reflects fairly intuitive ways in which the challenges presented by a particular dataset can be categorized, namely: large numbers of instances, large numbers of features, high class imbalance, and binary versus multiclass classification. 

\subsection{Model-Dataset Biclustering}\label{s:r2}
\begin{figure}
    \centering
    \includegraphics[width=\textwidth]{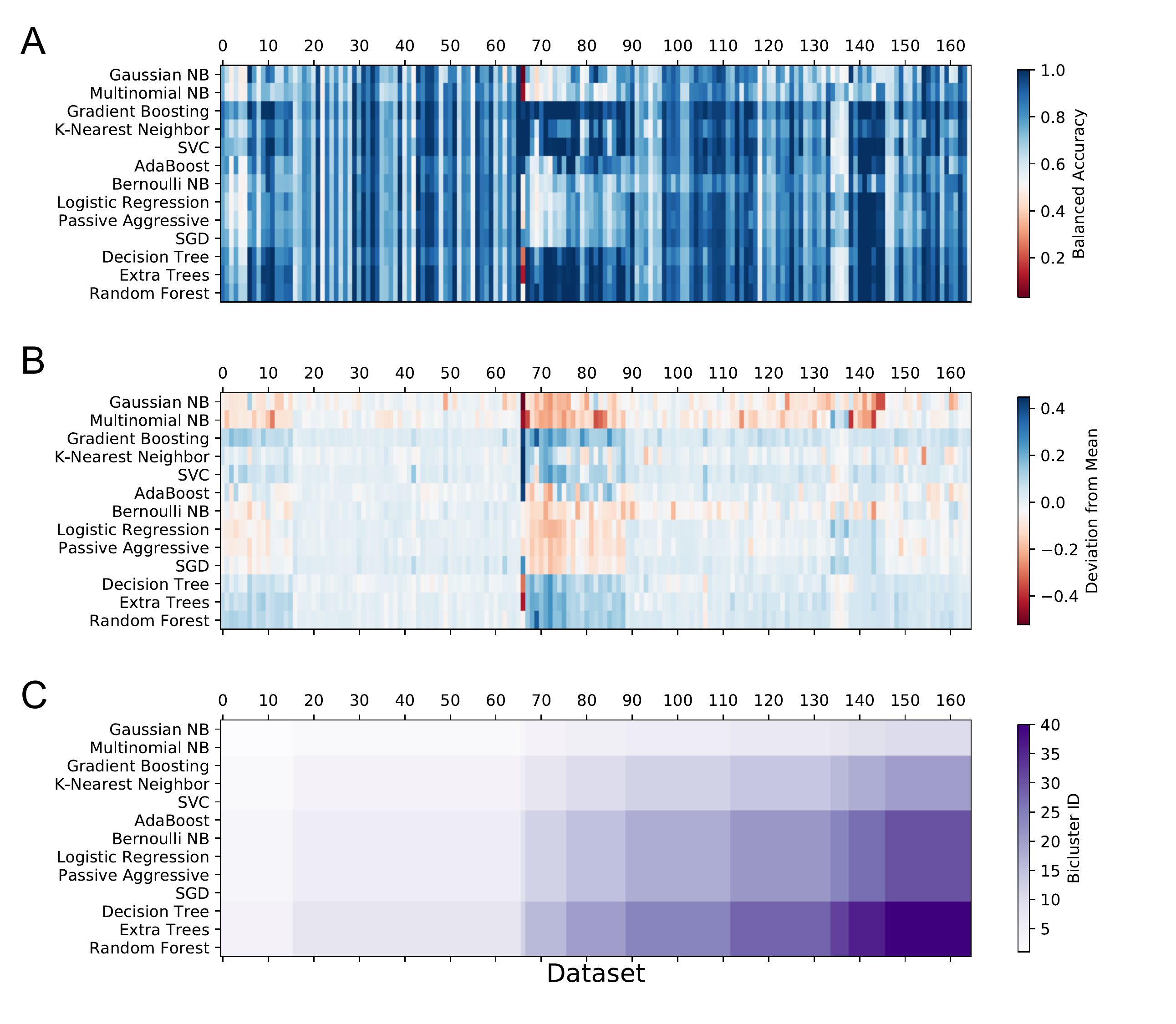}
    \caption{(A) Biclustering of the 13 ML models and 165 datasets according to the balanced accuracy of the models using their best parameter settings. (B) Deviation from the mean balanced accuracy across all 13 ML models.  Highlights datasets upon which all ML methods performed similarly versus those where certain ML methods performed better or worse than others. (C) Identifies the boundaries of the 40 contiguous biclusters identified based on the 4 ML-wise clusters by the 10 data-wise clusters.}
    \label{fig:bicluster}
\end{figure}

Figure \ref{fig:bicluster} summarizes the results of biclustering the balanced accuracy of the tuned models according to the ML method and dataset using a spectral biclustering algorithm~\cite{kluger2003spectral}.  The methods and datasets are grouped into 40 contiguous biclusters (4 ML-wise clusters by 10 data-wise clusters) in order to expose relationships between models and datasets.  Figure \ref{fig:bicluster}A presents the balanced accuracy. Figure \ref{fig:bicluster}B preserves the clustering from `A', but presents the deviation from the mean balanced accuracy among all 13 ML methods, in order to clearly identify datasets upon which all ML methods perform similarly, and those where some methods performed better than others.  Figure \ref{fig:bicluster}C simply delineates the 40 identified biclusters defined by balanced accuracy biclustering in Figure \ref{fig:bicluster}A and preserved in \ref{fig:bicluster}B.

It is interesting to note that the ML methods tend to group according to their underlying approach; for example, Gaussian and Multinomial Na\"{i}ve Bayes methods cluster together, Logistic Regression, Passive Aggressive and SGD cluster together (all hyperplane estimators), and the tree-based methods Decision Tree, Extra Trees and Random Forest also form a separate cluster.  Datasets that are derived from the same origin are observed to cluster in certain instances. For example, dataset cluster 1 (i.e. the left-most dataset cluster identified in the Figure \ref{fig:bicluster}C, including 4 separate biclusters) contains most of the \texttt{GAMETES} data sets; cluster 2 contains most of the \texttt{mfeats} datasets and the \texttt{Breast Cancer} datasets; and cluster 10 includes both of the \texttt{Wine Quality} datasets and several thyroid-related datasets (\texttt{new-thyroid}, \texttt{allhyper}, \texttt{allbp}, \texttt{allrep}). 

Figure~\ref{fig:bicluster}A allows us to interpret the utility of certain datasets in terms of difficulty across {\it all} methods and across {\it classes} of methods. For example, the light-blue stripes of low balanced accuracy indicate that none of the models achieve good performance on datasets 22, 118, and 164, which correspond to the \texttt{GAMETES Epistasis} datasets that are known to be difficult due to the lack of univariate correlations between features and classes and the high amount of noise. In contrast, nearly every method solves dataset 140 (\texttt{clean2}) with a high degree of accuracy because there are simple linear correlations between the features and classes and no noise.

Other clusters of datasets and ML methods reveal contrasts in performance. Dataset cluster 3 is the only cluster to contain a single dataset, the \texttt{parity5} problem, corresponding to dataset 66 in Figure~\ref{fig:bicluster}. This is a unique problem in which a ML method must be able to quantify whether the {\it number} of features with a given binary value is even or odd in order to correctly classify each instance. As a result, methods that consider the main effect of features independently are not able to solve it (e.g. the Na\"{i}ve Bayes methods). In contrast, methods with high capacity for interactions between features do well (e.g. Gradient Boosting, K-Nearest Neighbor, SVC). This contrast is also seen in cluster 4 (datasets 67 - 75), which contains several datasets with strong interactions between features (e.g. \texttt{tic-tac-toe}, \texttt{parity5+5}, and \texttt{multiplexer-6}). Again we observe a contrast between ML methods that make assumptions of linear independence and those that do not across this cluster of datasets. Contrasting Figure \ref{fig:bicluster}A with \ref{fig:bicluster}B helps to differentiate differences in overall performance on given datasets from differences in performance based on selected ML methodology. One important observation is that a reasonably large proportion of benchmarks included in this study yielded similar performance over the spectrum of ML methods applied.  This is likely because the signals identified in these datasets were either universally easy or difficult to detect. Furthermore, for those datasets where variable performance was observed, often a group of datasets clustered together with a similar signature of better than average/worse than average performance (see Figure \ref{fig:bicluster}B).

\begin{figure}
    \centering
    \includegraphics[width=0.95\textwidth]{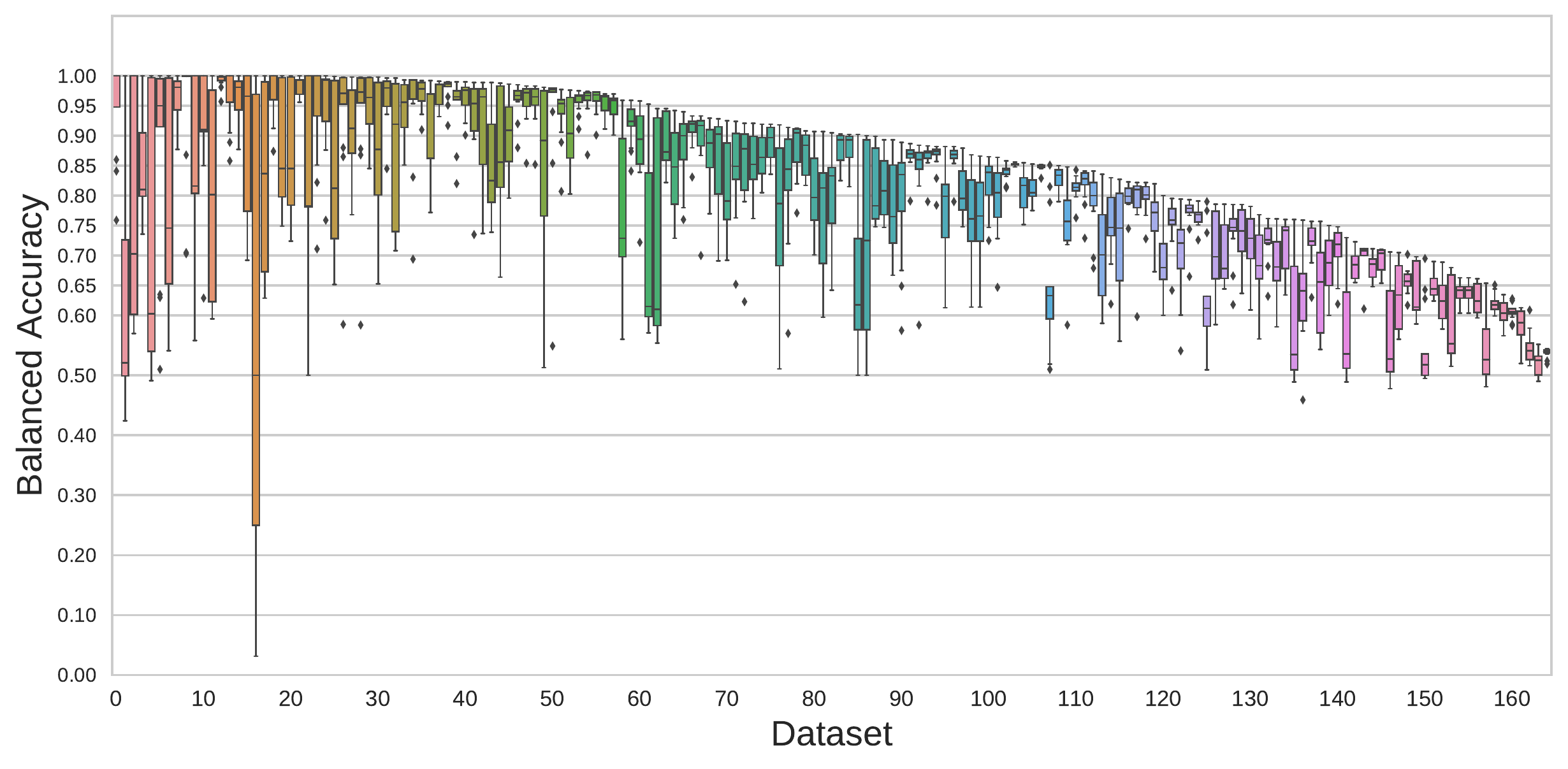}
    \caption{Accuracy of the tuned ML models on each dataset across the PMLB suite of problems, sorted by the maximum balanced accuracy obtained for that dataset.}
    \label{fig:tuned_acc}
\end{figure}

Overall, the current suite of datasets span a reasonable range of difficulty for the tested ML approaches. Figure~\ref{fig:tuned_acc} shows the distribution of scores for each tuned ML method for each dataset in the suite, sorted by best balanced accuracy score achieved by any method. The left-most dataset corresponds to \texttt{clean2}, mentioned above, and the right-most is \texttt{analcatdata\_dmft}, with a maximum accuracy score of 0.544 for the methods tested. Approximately half (87) of the current suite can be classified with a balanced accuracy of 0.9 or higher, and nearly all (98.8\%) of the datasets can be classified with a balanced accuracy of 0.6 or higher. Thus, although a range of model fidelity is observed, the datasets are biased towards problems that can be solved with a higher balanced accuracy.

\section{Discussion and Conclusion}

The primary goal of this paper is to introduce an ongoing research project for benchmarking ML methods.  Specifically, we have collected and curated 165 datasets from the most popular data repositories and introduced PMLB, a new evolving set of benchmark standards for comparing and evaluating different ML methods.  Apart from the repository itself, we have conducted a comprehensive analysis of the performance of numerous standard ML methods, which may be used as a baseline for evaluating and comparing newly developed ML methods, and assessed the diversity of these existing benchmark datasets to identify shortcomings to be addressed by the subsequent addition of further benchmarks in a future release.

Simplicity and diversity are the ultimate priorities of the PMLB suite. This motivated us to clean and standardize the presentation of datasets in the repository, develop a simple interface for fetching data, and include datasets from multiple sources. Interestingly, when we analyzed the meta-features of the datasets in PMLB, we found that most of the datasets fall into a handful of categories based on feature types, class imbalance, dimensionality and numbers of classes. We also found that by biclustering the performance of a set of different ML algorithms on the datasets, we could observe classes of problems and algorithms that work well or poorly in conjunction.   

Of course, PMLB is not yet a fully comprehensive benchmark suite for supervised classification methods. For instance, it currently excludes datasets with missing values or regression tasks and PMLB only has a handful of highly imbalanced datasets.  One approach to adding diversity, pursued by the KEEL repository, is to augment existing datasets to simulated missingness and noise.  However, we propose to avoid adding multiple variants of the same dataset, and instead identify and simulate new datasets with varying properties and meta-features to expand the PMLB suite and ``fill in the gaps'' of underrepresented problem types from a data science perspective.  As in the present study, we plan to use performance comparisons over a diversity of ML methods in order to identify a limited set of benchmark standards able to diversely identify methodological advantages and disadvantages.

We expect this future work to lead to a more comprehensive benchmark tool that will better aid researchers in discovering the strengths and weaknesses of ML methods, and ultimately lead to more thorough---and honest---comparisons between ML methods.


\acks{We thank Dr.~Andreas C.~M\"{u}ller for his valuable input during the development of this project. We also thank the Penn Medicine Academic Computing Services for the use of their computing resources. This work was supported by National Institutes of Health grants AI116794, DK112217, ES013508, EY022300, HL134015, LM009012, LM010098, LM011360, TR001263, and the Warren Center for Network and Data Science.}


\newpage

\appendix
\bibliography{references}

\end{document}